%% file: interspeech2022.tex
\documentclass[a4paper]{article}

\usepackage{INTERSPEECH2022}

\usepackage{latexsym}
\usepackage{hyperref}

\usepackage{changes}
\usepackage{lipsum}
\usepackage{enumitem}
\definechangesauthor[name={Danni}, color=violet]{DL}

\usepackage{booktabs}
\usepackage{multicol}
\usepackage{multirow}

\usepackage{amsmath}
\usepackage{algorithm}
\usepackage{algpseudocode}
\MakeRobust{\Call}
\usepackage{xurl}
\usepackage{subcaption}
\usepackage{graphicx}
\definecolor{myorange}{rgb}{1,0.5,0}
\definecolor{myblue}{RGB}{0,161,249}

\title{From Start to Finish: Latency Reduction Strategies for Incremental Speech Synthesis in Simultaneous Speech-to-Speech Translation}
\name{Danni Liu$^1$, Changhan Wang$^2$, Hongyu Gong$^2$, Xutai Ma$^{2,3}$, Yun Tang$^2$, Juan Pino$^2$}
\address{
        $^1$Maastricht University, The Netherlands
        $^2$Meta AI, USA
        $^3$Johns Hopkins University, USA}
\email{{\footnotesize $^1$danni.liu@maastrichtuniversity.nl,$^2$\{changhan,hygong,yuntang,juancarabina\}@fb.com,$^3${xutai\_ma@jhu.edu}}}
\begin{document}
\maketitle

\begin{abstract}
Speech-to-speech translation (S2ST) converts input speech to speech in another language.
A challenge of delivering S2ST in real time is the accumulated delay between the translation and speech synthesis modules.
While recently incremental text-to-speech (iTTS) models have shown large quality improvements,
they typically require additional future text inputs to reach optimal performance.
In this work, we minimize the initial waiting time of iTTS by adapting the upstream speech translator to generate high-quality pseudo lookahead for the speech synthesizer.
After mitigating the initial delay, we demonstrate that the duration of synthesized speech also plays a crucial role on latency.
We formalize this as a latency metric and then present a simple yet effective duration-scaling approach for latency reduction.
Our approaches consistently reduce latency by 0.2$-$0.5 second without sacrificing speech translation quality.\footnote{\url{https://github.com/pytorch/fairseq/pull/4184}}
\end{abstract}

\section{Introduction}
Speech-to-speech translation (S2ST)
\cite{lavie2017,nakamura2006,sudoh} is the task of translating input speech utterances into speech in another language.
Compared to translating into text alone, delivering translation into speech offers users with increased accessibility.
While recent end-to-end S2ST methods \cite{tjandra2019,translatoron,s2ut,translatoron2} have shown encouraging results,
the pipeline approach based on intermediate text representations remains a strong baseline, which currently has not yet been surpassed by its end-to-end counterparts \cite{translatoron,s2ut,translatoron2}. 
This motivates the relevance of bringing this approach under the more challenging condition of \textit{simultaneous} S2ST. 
To enable real-time communication,
all components of the S2ST pipeline must be optimized for incremental inference.
Here we consider a direct speech-to-text (ST) translation and a text-to-speech (TTS) component.
For the ST module, handling the complex mappings from acoustic to textual representations often requires large model size, which comes with high inference computation.
Moreover, to account for word reordering in translation, the ST module often requires additional input context \cite{ma-etal-2019-stacl,ma-etal-2020-simulmt} before outputting the translated text.
Given the well-justified computational and algorithmic delays of the upstream simultaneous ST task,
the incremental TTS (iTTS) module must be \textit{lightweight} and \textit{minimal-latency}.
For latency reduction in iTTS, current approaches \cite{ma-etal-2020-incremental,stephenson21_interspeech,pseudolookahead,pseudolookaheaddistilled} improve the system response speed by minimizing the amount of future input context (or \textit{lookahead}) required by stand-alone TTS systems.
In this work, we show that S2ST presents a unique opportunity to circumvent the initial delay from lookahead: we leverage the access to input speech for latency reduction.
Moreover, we show that faster system response, or in other words latency reduction at the \textit{start}, does not guarantee low latency when \textit{finishing}.
Due to the non-overlapping constraints of incremental TTS (i.e., no upcoming audio can be played when the current word is still playing),
when output speech has stretched duration, the final latency is substantially increased.
We explicitly account for this constraint in our framework.
Our contributions are:
\begin{itemize}[leftmargin=*]
\itemsep0em 
\item
We propose a lightweight input-speech-guided pseudo lookahead mechanism that reduces the \textit{starting} latency of iTTS.
\item
We show that the \textit{duration} of synthesized speech plays a crucial role in \textit{final} latency, and provide a simple yet effective duration-scaling approach that leverages recent advances in non-autoregressive TTS.
\item 
We present an improved iTTS model that differentiates partial and full inputs, which,
in combination with the above techniques, consistently reduces latency by 0.2$-$0.5$s$ while retaining speech translation quality.
\end{itemize}

\input{figures_interspeech/overview}
\vspace*{-3mm}
\section{Related Work}
\input{sections/related_work_interspeech}

\section{Approach}

\subsection{Input-Speech-Guided Pseudo Lookahead}
\label{subsec:pseudolookahead}
Incremental TTS quality is often improved when allowing a lookahead to some future input words  \cite{ma-etal-2020-incremental,lookahead,stephenson21_interspeech}, which on the other hand causes delay.
Several works \cite{stephenson21_interspeech,pseudolookahead,pseudolookaheaddistilled} use pretrained language models (LMs) to generate pseudo lookahead.
While this improves speech quality to some extent, the LM-predicted pseudo lookahead tends to disagree with the ground-truth lookahead \cite{stephenson21_interspeech,pseudolookahead}.
It has further been shown that only the \textit{correct} pseudo lookahead improves output prosodic features \cite{stephenson21_interspeech}.
Our initial experiments on controlled replacements of ground-truth lookaheads also confirms this finding.

While pretrained LMs are powerful at many generation tasks, 
generating the exact next words for TTS is challenging, especially at the beginning of the sentence, where the model can only predict likely words in the \textit{average case}. 
In S2ST, however, we can additionally leverage the contextual information from the source speech.
With a separate LM, assuming greedy decoding for illustrative purposes, the pseudo lookahead $w_t$ at time step $t$ would be
${\mathrm{argmax}_w}\,P(w_t | w_{1, \dots t-1})$,
which is only conditioned on the currently available words  $w_{1, \dots, t-1}$.
In contrast, the pseudo lookahead generated from the upstream ST system is additionally conditioned on the currently-available source speech $\textbf{s}$:
${\mathrm{argmax}_w}\,P(w_t | w_{1, \dots t-1}, \textbf{s}).$
We hypothesize the source speech can help us generate more accurate pseudo lookahead, and therefore minimize the performance gap to the case of using ground-truth lookahead.
\autoref{fig:approach} illustrates our idea in comparison to existing methods.

To generate the pseudo lookahead, 
for every token produced by the upstream ST system, 
we trigger an additional decoding step.
At the next timestep, the decoder continues at the hidden state cached before the pseudo lookahead is generated.
This mechanism ensures that the actual ST outputs remain unchanged regardless of pseudo lookahead generation. 
We note the procedure also applies to multiple pseudo lookahead steps.

Compared to separate LMs, the advantage of our approach is two-fold: 
First, it is easier to deploy as no extra component is introduced into the translation pipeline. 
Moreover, from a modeling point of view, this approach has access to the input speech, which likely contains more semantic information on the text to synthesize. 
We provide more analysis on this in \S \ref{subsec:combine_st}.

\subsection{Speech-Duration-Aware Latency Metric} \label{subsec:latency}
When evaluating iTTS latency, existing  works \cite{stephenson21_interspeech,neuraliTTS, ma-etal-2020-incremental, RLiTTS}
mainly 
focus on the amount of input text the model requires before starting to produce output audio, e.g. waiting for $X$ words.
While this does correspond to the initial system response time, it does not fully account for the real-time constraints in S2ST applications that often translate \textit{consecutive} utterances.
As illustrated in \autoref{subfig:example_1},
although the TTS module starts with minimal delay immediately after receiving the first text token, 
the full utterance still finishes comparatively late
because the output speech has prolonged duration.
Since multiple output words cannot play concurrently, output audios with constantly long durations
with can congest the system and cause an ever-increasing latency.
In comparison, in \autoref{subfig:example_2}, although the model initially waits for an additional word, its final latency is lower thanks to the time saved from shorter output durations (i.e. speaking faster).
While admittedly the speaking speed of synthesized speech cannot be considered alone without naturalness and intelligibility,
this example showcases the crucial impact of duration on latency.

Motivated by this observation, we adopt a latency metric that account for the time elapsed between end points of the input and output illustrated in \autoref{fig:latency}.
Algorithm \ref{alg:latency} shows how we measure the latency for an utterance.
It depends on three factors: 
1) when the input for synthesizing a given word arrived (\textsc{emit\_time}$(\cdot)$), 
2) computation time for synthesizing the word (\textsc{compute\_time}$(\cdot)$),
3) when the previous word finished playing (dependent on \textsc{duration}$(\cdot)$).
As the texts from a simultaneous ST system to the TTS module are time-stamped, \textsc{emit\_time}$(\cdot)$ can be easily derived.

\input{figures/duration}
\input{tables_interspeech/latency_calc}

\subsection{Improving Incremental TTS Module} \label{subsec:prefix_training}
Having demonstrated the impact of output speech duration on latency, we need a TTS system that enables direct duration control.
Among current TTS models, non-autoregressive (NAR) ones are suitable candidates, as the source-target alignment (therefore the duration of phonetic units) come from a dedicated duration predictor.
In contrast, with autoregressive models (e.g. \cite{tacotron,tacotron2}), the output speech duration is a by-product of source-target attention and is less straightforward to control.
Therefore, we build upon a widely-used NAR TTS model, FastSpeech 2 \cite{fastspeech2}.
Given $j$ current phonemes and $k$ (pseudo) lookahead phonemes, we synthesize up till $X_{j+k}$ but only output those frames up to $X_j$.
We use the duration predictor to derive word boundaries in the synthesized output, as also done in \cite{stephenson21_interspeech}.

As the TTS module consumes partial inputs at inference time, we augment the training set by prefixes.
The prefix augmentation is similar to the approach by \cite{neuraliTTS}, except that they use shorter subsets that are not necessarily prefixes\footnote{Our preliminary experiments with this approach did not show consistent improvements as prefix augmentation.}.
Moreover, as prosodic features tend to differ during and at the end of sentences \cite{cooper1981segmental,berkovits1993utterance}, 
we use the presence of the end-of-sentence (EOS) token to distinguish between partial and full sentences: Full input sequences are followed by the EOS token while partial sequences are not.
We expect the presence of EOS token to signal the model to synthesize the suitable prosody at test time.

\section{Experiments} \label{subsec:exp_setup}
\subsection{Data}
We build S2ST systems for Spanish to English (es$\rightarrow$en) and vice versa (en$\rightarrow$es). 
\autoref{tab:stats} shows the dataset statistics.

{
\setlength{\parindent}{0cm}
\textbf{ST Data}
For es$\rightarrow$en, the ST system is trained on the Fisher Spanish-English \cite{fisher} dataset.
For en$\rightarrow$es we use MuST-C v1.0 \cite{di-gangi-etal-2019-must} and report performance on \texttt{tst-COMMON}.
For audio inputs, we extract 80-dimensional filterbank 
with global ceptral mean variance normalization.
For text, we learn byte pair encoding \cite{sennrich-etal-2016-neural} using SentencePiece \cite{kudo-richardson-2018-sentencepiece} with size 500 and 10$k$ for Fisher and MuST-C respectively.
For Fisher, the vocabulary size is intentionally kept small due to the limited corpus size.
}

{
\setlength{\parindent}{0cm}
\textbf{TTS Data}
We train incremental TTS models for English and Spanish on
LJSpeech \cite{ljspeech17} and CSS10 Spanish \cite{css10} respectively. 
The phoneme durations are extracted by Montreal Forced Aligner (MFA) \cite{mfa} using pretrained acoustic models in the MFA toolkit.
The phoneme representations are stressed-ARPAbet and GlobalPhone \cite{schultz-schlippe-2014-globalphone} respectively. 
}
\input{tables_interspeech/dataset}

\subsection{Models and Training} \label{subsec:training}
The simultaneous ST system is a Transformer model with a wait-$k$ \cite{ma-etal-2019-stacl} policy.
Following \cite{ma-etal-2020-simulmt}, we use a fixed pre-decision of 7 encoder states where each state is an audio frame of 40$ms$. 
As a result, a text token is generated every 0.28$s$.
The model architecture is the \texttt{s2t\_transformer\_s} model from the \textsc{fairseq S2T} toolkit \cite{wang-etal-2020-fairseq}. 
For TTS, we follow the hyperparameters from~\cite{fastspeech2} and train for 200$k$ updates.
We train HiFi-GAN with configuration $V$3 \cite{hifigan}.
When applying prefix augmentation, we keep an equal ratio between prefixes and full sentences.
The prefixes are 1/3 or 2/3 of the full sentence lengths.

\subsection{Evaluation}
\textbf{Automatic quality evaluation.}
We concatenate the incremental speech chunks into full utterances for evaluation.
Following \cite{Wave-Tacotron,wang-etal-2021-fairseq}, we use mean Mel ceptral distortion (MCD) \cite{MCD} and character error rates (CER) to evaluate TTS quality. 
MCD is computed between the HiFi-GAN vocoded utterances from the ground-truth and synthesized Mel spectrograms.
The English ASR system is a wav2vec 2.0 \cite{wav2vec} model.
The Spanish ASR system is an XLSR-53 \cite{xlsr53} model fine-tuned on Common Voice \cite{ardila-etal-2020-common} Spanish data\footnote{\url{https://huggingface.co/jonatasgrosman/wav2vec2-large-xlsr-53-spanish}}.
For speech-to-speech experiments, we transcribe the synthesized utterances with the corresponding ASR system and evaluate BLEU\footnote{sacrebleu \cite{post-2018-call} ID: BLEU+case.mixed+numrefs.1\\+smooth.exp+tok.13a+version.1.5.1} against the reference translation.
As the ASR outputs are uncased and unpunctuated, we normalize the reference text accordingly before evaluating BLEU.

{
\setlength{\parindent}{0cm}
\textbf{Subjective quality evaluation.}
As human evaluation of S2ST currently still lacks standardized procedures, we only conduct subjective evaluation of TTS quality. 
We collect mean opinion scores (MOS) on MTurk, where annotators rate speech naturalness on a Likert scale.
We evaluate the first 100 utterances of the LJSpeech test set, and the full test set of CSS10.
Each utterance receives 15 ratings from different annotators.

\textbf{Latency evaluation.}
We measure latency using the metric from \S \ref{subsec:latency}.
The utterances are synthesized one at a time on an Nvidia V100 GPU.
Computation time is included in the latency evaluation.
When evaluating the TTS module alone, as we do not directly have timestamps of the input words, we use timestamps derived by forced alignment using MFA \cite{mfa}.
}

\input{tables_interspeech/main_res}
\input{tables_interspeech/s2st}

\subsection{Incremental TTS Results} \label{subsec:iTTS}
We first evaluate the incremental TTS module alone (\autoref{tab:main_res}).
Results of the improved model are in lines $(6$-$8)$ and $(14$-$16)$.

{
\setlength{\parindent}{0cm}
\textbf{Impact of output duration on latency.}
In lines $(4)$ and $(12)$,
we see that naively decoding substantially degrades quality.
Interestingly, although this approach does not wait for any future inputs, its latency is still much higher than other strategies that involve more waiting (lines $(5)$ and $(13)$).
By manual inspection, we confirm the increased latency is caused by prolonged durations of the synthesized audios\footnote{The long-duration phenomenon is also reported in \cite{stephenson21_interspeech}.}.
In particular, the average length of the synthesized audios is 5.9 and 7.7$s$ for lines $(5)$ and $(4)$ respectively, 
while the ground truth audio length is 6.0$s$ on average. 
As incorporating lookahead leads to more suitable durations, we achieve better latency despite the initial waiting.
This phenomenon corresponds to the example in \autoref{fig:latency}, where less lookahead does not necessarily reduce latency.

\textbf{Improved iTTS module.}
\autoref{tab:main_res} shows the prefix-augmented model outperforms the baseline under all experimented decoding strategies.
One reason for the gain is the distinguishing of partial and full inputs (\S \ref{subsec:prefix_training}).
With the baseline model, every partial sentence would be synthesized as if it were a full sentence, 
therefore carrying end-of-sentence prosodic features such as duration stretching \cite{cooper1981segmental}} and dropping intonations (confirmed by manual inspections).
By explicitly indicating full inputs with the EOS tags, we ease the prosody prediction task.
Given the favorable results in terms of both quality and latency, we use the prefix-augmented iTTS system in the next S2ST experiments.

\subsection{Simultaneous Speech-to-Speech Translation Results} \label{subsec:combine_st}
In \autoref{tab:s2t}, we report the performance of S2ST systems of different latency regimes.
We include the BLEU score of recent offline systems \cite{s2ut,ma2022direct} as quality upper bounds.
For each upstream ST model, we compare the performance of the following iTTS decoding strategies: 
1) \textbf{Lookahead 1 word}: wait for the next word from the ST system; 
2) \textbf{No lookahead}: directly synthesize currently available words without waiting; 
3) \textbf{+Pseudo}: directly synthesize without waiting, additionally use pseudo lookahead generated by upstream ST system.

{
\setlength{\parindent}{0cm}
\textbf{S2ST quality-latency tradeoff.}
In \autoref{tab:s2t}, by contrasting the first two decoding strategies (``lookahead 1 word'' vs ``no lookahead''), we clearly observe the quality-latency tradeoff:
Without the lookahead, we consistently achieve lower latency of 0.2 to 0.5$s$.
This comes with lower output quality, 
as shown by the loss of $\sim$1.5 BLEU on en$\rightarrow$es and $\sim$1.0 BLEU on es$\rightarrow$en.
By incorporating our proposed ST pseudo lookahead, we are able to match the output quality when using the actual lookahead (``lookahead 1 word''), while avoiding the delay.
These findings suggest that our ST-generated pseudo lookahead achieves more efficient quality-latency tradeoff than existing approaches.
Moreover, to the best of our knowledge, this is the first evidence that pseudo lookahead matches the performance of ground-truth lookahead in term of output speech quality.

\textbf{Importance of accurate pseudo lookahead.}
To confirm the gains indeed come from high-quality pseudo lookahead,
we evaluate the accuracy of the ST-generated pseudo lookahead, i.e. how many pseudo lookahead tokens agree with the word generated in the next time step.
The accuracy is 70\%, 76\%, 81\%, 83\% for Wait-$k=3,5,7,9$ on Fisher es$\rightarrow$en.
Our initial experiments with separate LMs yielded low accuracy under 20\% and did not improve output speech quality.
In \autoref{tab:analysis_pseudolookahead},
we report BLEU scores of random pseudo lookahead sampled from the ST model's vocabulary.
The consistent degradation (up to 3.1 BLEU vs ``ST pseudo lookahead" and 1.9 BLEU vs ``no lookahead") indicates that the pseudo lookahead must be accurate to improve synthesis quality.
Low-accuracy pseudo lookahead is more harmful than not using any lookahead.

\begin{table}[h]
\caption{BLEU$(\uparrow)$ scores of different lookahead strategies for speech-to-speech translation experiments on es$\rightarrow$en.
}
	\centering
	\setlength\tabcolsep{5.5pt} 
	\begin{tabular}{lccccc}
		\toprule
		\shortstack[c]{\textbf{Wait-$k$}}  & 
        \shortstack[c]{\textbf{$k=3$}}  & 
        \shortstack[c]{\textbf{$k=5$}}  &
        \shortstack[c]{\textbf{$k=7$}} & 
        \shortstack[c]{\textbf{$k=9$}} \\ 
		\midrule
		Lookahead 1 word &  
		27.5 & 31.5 & 32.7 & 34.5 \\
		No lookahead &  
		26.0 & 29.8 & 31.1 & 33.0 \\
		ST pseudo lookahead &  
		27.1 & 31.3 & 32.5 & 34.3 \\
		Random lookahead    &  
		24.4 & 28.4 & 28.8 & 31.1 \\
		\bottomrule
	\end{tabular}
	\label{tab:analysis_pseudolookahead}
\end{table}

\subsection{Output Duration Control for Latency Reduction} \label{subsec:speed}
Our upstream wait-$k$ ST system generates tokens at a fixed rate of 0.28s (\S \ref{subsec:training}). 
We now study the impact on latency when the tokens come in at a faster rate.
As shown in upmost line in \autoref{fig:duration}, a minor difference of 0.06$s$ per token (0.28$\rightarrow$0.22$s$) increases latency by 1.3 seconds, as a result of accumulated latency (c.f. Figure \ref{subfig:example_1}).
Therefore,
we speed up the output utterance and investigate its tradeoff with quality.
Recall from \S \ref{subsec:prefix_training} that the duration prediction module enables us to scale the output duration without altering other prosodic features.
As shown by the bottom two lines in \autoref{fig:duration}, when moderately speeding up the output speech by scaling the durations by 0.95 or 0.9, we can counteract the latency induced by faster incoming text tokens.
From the quality metrics, this comes with a minimal impact on audio intelligibility and naturalness, with neither CER and MOS differing noticeably 
from the original values of 5.1\% and 3.77.
These findings showcase the effectiveness of our iTTS module in controlling duration for latency reduction.

\begin{figure}[ht!]
    \centering
    \includegraphics[
    trim={0.2cm 0.5cm 0.2cm 0.4cm},clip,width=0.95\linewidth]{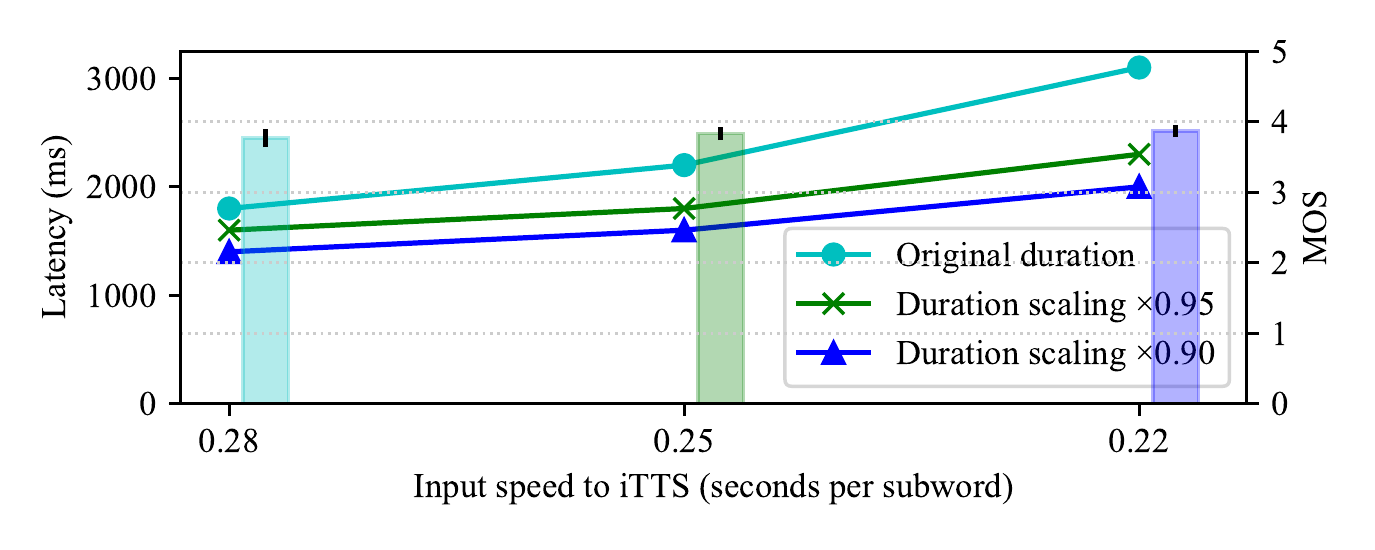}
\caption{\label{fig:duration} 
S2ST latency is sensitive to the speed of incoming text tokens to iTTS.
By duration scaling (\S \ref{subsec:prefix_training}), we reduce latency by $\sim$ 0.7$s$ without harming output naturalness measured by MOS.
}
\end{figure}

\vspace*{-5mm}
\section{Conclusion}
In this work, we improve speech-to-speech translation pipelines for simultaneous speech-to-speech translation.
We adapt the upstream speech-to-text translation system to generate high-quality pseudo lookahead for the TTS module,
allowing the latter to deliver output translation with minimal delay.
After demonstrating the crucial role of duration on latency, we 
build on recent non-autoregressive models for duration-controllable TTS.
We further show our prefix augmentation procedure substantially improves the incremental speech quality.

\bibliographystyle{IEEEtran}
\bibliography{interspeech2022} 

\end{document}

%% file: figures_interspeech/overview.tex
\begin{figure}[t] 
    \centering
    \includegraphics[
    trim={0 0cm 0 0},clip,width=0.73\linewidth]{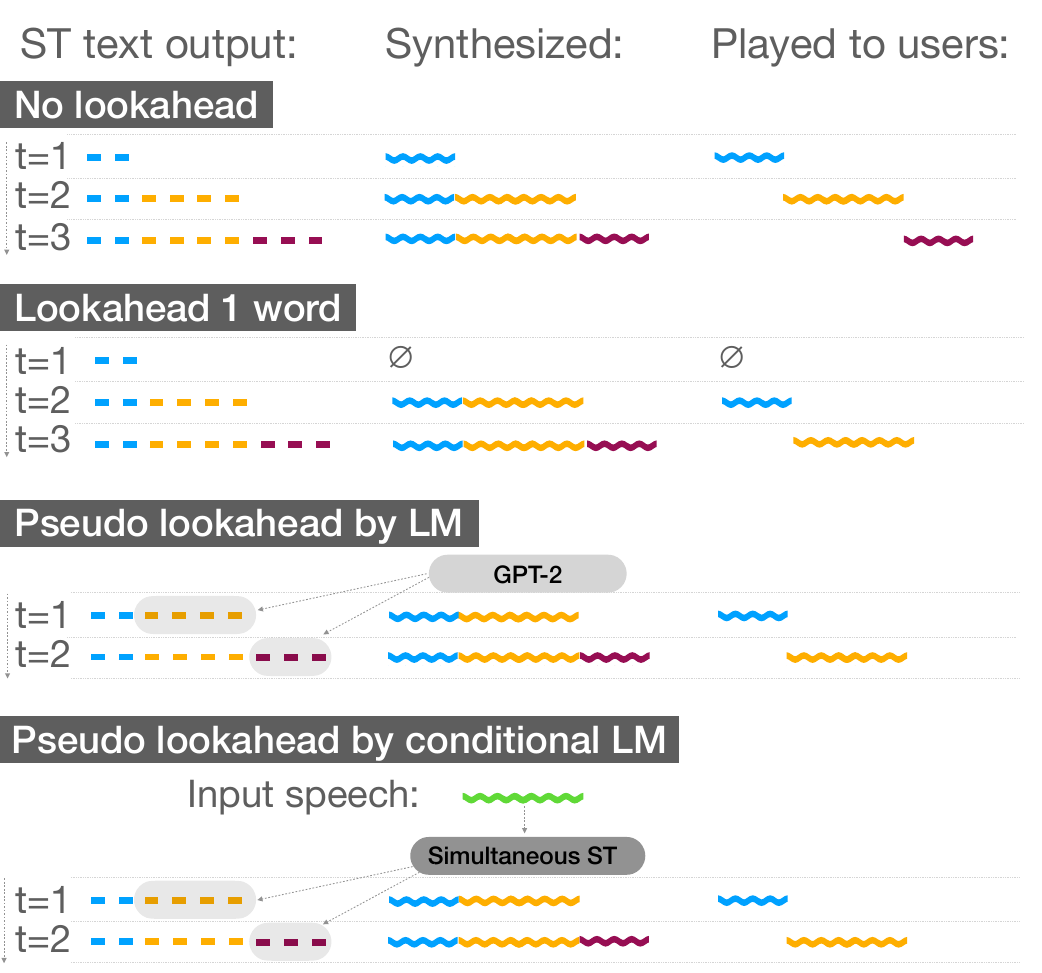}
\caption{\label{fig:approach} Example of synthesizing \{{\color{myblue} $w_1$}, {\color{myorange} $w_2$}, {\color{violet} $w_3$}\}.
The initial delay caused by the lookahead could be mitigated by pseudo lookahead generated by monolingual LMs \cite{stephenson21_interspeech,pseudolookahead}.
We instead adapt the upstream ST system to generate pseudo lookahead.
}
\end{figure}

%% file: sections/related_work_interspeech.tex
Initial works on iTTS \cite{neuraliTTS,ma-etal-2020-incremental,lookahead,RLiTTS} build upon autoregressive models, notably Tacotron \cite{tacotron,tacotron2}, 
and recently \cite{stephenson21_interspeech} extends the study to non-autoregressive models.
An often-discussed topic in iTTS is the role of \textit{lookahead}:
\cite{ma-etal-2020-incremental} use a lookahead \cite{ma-etal-2019-stacl} for both the text-to-spectrogram model and vocoder, and achieve performance on par with offline systems.
\cite{lookahead} show that shorter words need more lookahead.
\cite{stephenson21_interspeech,pseudolookahead,pseudolookaheaddistilled} utilize GPT-2 \cite{radford2019language} to generate pseudo lookahead for iTTS.
Since the lookahead causes initial delays for the iTTS module, current works mainly focus on minimizing its amount.
Meanwhile, the relation between the lookahead and latency when \textit{finishing} is still under-explored.
Moreover, to the best of our knowledge, no previous work has leveraged source speech to generate lookahead for iTTS.


%% file: figures/duration.tex
\begin{figure}[ht!]
\centering
\subfloat[Start early, finish late]{%
  \includegraphics[trim={0.3cm 5.1cm 2cm 0},clip,width=0.45\linewidth]{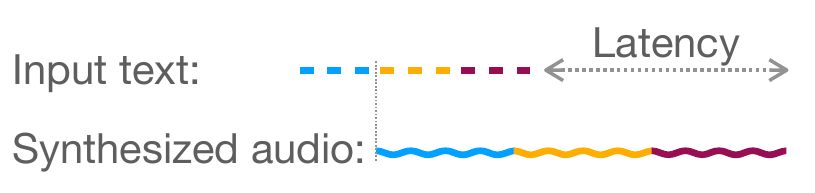}%
  \label{subfig:example_1}%
}\qquad
\subfloat[Start late, finish early]{%
  \includegraphics[trim={0.3cm 5cm 2.2cm 0.2cm},clip,width=0.42\linewidth]{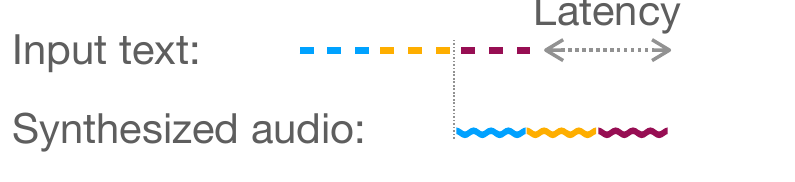}%
  \label{subfig:example_2}%
}
\caption{\label{fig:latency} Example of incrementally synthesizing words {\color{myblue} $w_1$}, {\color{myorange} $w_2$}, {\color{violet} $w_3$} and the impact of output duration on latency. 
Lengths of wavy lines indicate duration of synthesized speech.
}
\end{figure}

%% file: tables_interspeech/latency_calc.tex
\begin{algorithm}[ht!]
    \captionsetup{labelfont={sc,bf}} 
	\caption{Utterance-level latency for synthesizing a sentence with tokens \{$w_1,\cdots,w_n$\}}
\begin{algorithmic}[1]
	\Procedure{CalculateLatency}{$\{w_1, \cdots, w_n\}$}
		\State $t \gets 0$ 
		\For {$w_i \in \{w_1, \cdots, w_n\}$}					
		    \State $s \gets \Call{emit\_time}{w_i}$ 
			\State $p \gets \Call{max}{t, s + \Call{compute\_time}{w_i}}$   \Comment{{\color{darkgray}Synthesized audio can only play if: 1) previous word finished playing; 2) current word has been synthesized}}
			\State $t \gets p + \Call{duration}{w_i}$  
		\EndFor
		\State \Return $t - \Call{emit\_timestamp}{w_n}$ 
	\EndProcedure
\end{algorithmic}
\label{alg:latency}
\end{algorithm}

%% file: tables_interspeech/dataset.tex
\begin{table}[ht!]
\caption{\label{tab:stats}Dataset statistics.}
\setlength\tabcolsep{4pt} 
\centering
\begin{tabular}{ llccrrrrrr } 
\toprule
\multirow{1}{*}{\shortstack[c]{ \\\\ \textbf{Task} }}  
& 
\multirow{1}{*}{\shortstack[c]{ \\\\ \textbf{Corpus} }} 
&
\multirow{1}{*}{\shortstack[c]{ \\\\ \textbf{Lang.} }} 

          & \multicolumn{2}{c}{\textbf{Train}} && \multicolumn{2}{c}{\textbf{Test}}\\
          \cmidrule{4-5} \cmidrule{7-8}
&&&
\# sent. &
hrs &&
\# sent. &
hrs \\
\midrule
ST &
Fisher &
es$\rightarrow$en &
126$k$ & 162 
&& 3,641 & 4.5
\\
& MuST-C & 
en$\rightarrow$es &
270$k$ & 504    
&& 2,052 &
4.2 \\
\midrule
TTS &
LJSpeech & 
en & 
12$k$
&22&
&
523&
1.0\\
&
CSS10 &
es &
11$k$&
23&
&
107 &
0.2\\
\bottomrule
\end{tabular}
\end{table}

%% file: tables_interspeech/main_res.tex
\begin{table}[t!]
\caption{\label{tab:main_res} iTTS performance.
We report 95\% confidence interval (CI) over 3 runs for CER, and over valid ratings for MOS.
CI for MCD and latency is $<$0.1 and omitted for brevity.}
\centering
\setlength\tabcolsep{0.1pt} 
\begin{tabular}{ llccccccrc } 
\toprule
& \multirow{1}{*}{\shortstack[c]{ \\\\\\ \textbf{Condition} }}  
          & \multicolumn{3}{c}{\textbf{Quality}} && \multicolumn{1}{c}{\textbf{Latency}}\\
            \cmidrule{3-5}  
&& MCD$\downarrow$ & 
CER(\%)$\downarrow$ &
\shortstack[c]{MOS}$\uparrow$ &&
(seconds) \\ 
\midrule
\multicolumn{2}{l}{\textbf{LJSpeech (English)}} \\
$(1)$ &
Original audio & - &  3.3$\pm$0.0 &  4.16$\pm$0.15 &    & - \\
$(2)$ &
 Ground truth Mel   & - & 3.2$\pm$0.0 & 3.73$\pm$0.13  &    & - \\
\midrule
$(3)$ &
 Offline
 & 3.5  & 4.8$\pm$0.1 & 3.72$\pm$0.14 &    & 6.2  \\
$(4)$ &
 No lookahead
 & 4.7 & 8.2$\pm$0.2  & 3.02$\pm$0.18  & & 2.4 \\
$(5)$ &
 Lookahead 1 word
 & 3.9 & 5.6$\pm$0.2 & 3.42$\pm$0.16 & & 1.4 \\
\midrule
& \multicolumn{2}{l}{\textbf{After augmentation:}}  \\
$(6)$ &
 Offline
 & 3.6 & 5.0$\pm$0.2 & 3.85$\pm$0.15 &   & 6.5 \\
$(7)$ &
%
 No lookahead
 & 3.8 & 5.9$\pm$0.3 & 3.13$\pm$0.18 & & 1.1 \\
$(8)$ &
 Lookahead 1 word
%
 
 & 3.7 & 5.1$\pm$0.1 & 3.77$\pm$0.13 & &  1.8 \\
\midrule
\multicolumn{2}{l}{\textbf{CSS10-es (Spanish)}} \\
$(9)$ &
 Original audio     & - &   4.9$\pm$0.0 & 4.78$\pm$0.05  &  & - \\
$(10)$ &
 Ground truth Mel   & - & 5.2$\pm$0.0           & 4.56$\pm$0.07 &  & - \\
\midrule
$(11)$ &
 Offline
 & 3.6 & 5.0$\pm$0.3 & 4.34$\pm$0.08 &  & 6.4 \\
$(12)$ &
 No lookahead
 & 5.1  & 13.2$\pm$0.5 & 1.48$\pm$0.08 &  & 2.1 \\
$(13)$ &
 Lookahead 1 word
 & 3.9 & 8.5$\pm$0.9 & 3.04$\pm$0.12 &  & 1.5 \\
\midrule
& \multicolumn{2}{l}{\textbf{After augmentation:}}  \\
$(14)$ &
 Offline
 & 3.6 & 5.0$\pm$0.3 & 4.45$\pm$0.07 &  & 6.4 \\
$(15)$ &
 No lookahead
 & 3.8 & 7.1$\pm$0.5 & 3.11$\pm$0.13 &  & 0.9 \\
$(16)$ &
 Lookahead 1 word
 & 3.6 & 5.2$\pm$0.2 & 3.95$\pm$0.10 &  & 1.5 \\
\bottomrule

\end{tabular}
\end{table}

%% file: tables_interspeech/s2st.tex
\begin{table}[ht]
\caption{S2ST quality for es$\rightarrow$en (Fisher) and en$\rightarrow$es (MuST-C) in upper and lower sections.
When using pseudo lookahead from the ST system, we achieve output speech quality on par with ground-truth lookahead without waiting for an extra word.
}
\setlength\tabcolsep{0.6pt} 
\centering
\begin{tabular}{lcccccccccccc}
	\toprule
	&
	\multirow{1}{*}{\shortstack[c]{ \\\\\\\\ \textbf{Wait}-$k$ }}  &
	\multicolumn{8}{c}{\textbf{iTTS Decoding Strategy}} \\
	&&
	\multicolumn{2}{c}{Lookahead 1 word} &&
	\multicolumn{2}{c}{No lookahead} &&
	\multicolumn{2}{c}{$+$Pseudo} \\
	\cmidrule{3-4} \cmidrule{6-7} \cmidrule{10-11} 
	&
	& latency($s$)$\downarrow$ & BLEU$\uparrow$  &
	& latency($s$)$\downarrow$ & BLEU$\uparrow$  &
	& BLEU$\uparrow$  \\

	\midrule
	& $3$  & 1.3 & 27.5 && 0.9 & 26.0 && 27.1\\ 
	& $5$  &  1.4 & 31.5  && 1.1 & 29.8 && 31.3\\ 
	& $7$  &  1.6 & 32.7  && 1.3  & 31.1 && 32.5\\ 
	& $9$  &  1.7 & 34.5  && 1.5 & 33.0  && 34.3\\ 
    \cmidrule{3-9}
	& \multicolumn{1}{l}{Offline \cite{s2ut}} & \multicolumn{8}{c}{39.5} \\
	\midrule
	& $3$ & 2.3 & 14.4 && 1.8 & 13.5  && 14.4 \\
	& $5$ & 2.5 & 17.1 && 2.1  & 16.1 && 17.1 \\
	& $7$ & 2.8  & 18.7  && 2.4 & 18.5 && 18.8 \\
	& $9$ & 2.9  & 19.4 && 2.5 & 18.5 && 19.5 \\
	\cmidrule{3-9}
	& \multicolumn{1}{l}{Offline \cite{ma2022direct}} & \multicolumn{8}{c}{24.4} \\
	\bottomrule
\end{tabular}
\label{tab:s2t}
\end{table}

%% file: interspeech2022.bbl
\begin{thebibliography}{10}
\providecommand{\url}[1]{#1}
\csname url@samestyle\endcsname
\providecommand{\newblock}{\relax}
\providecommand{\bibinfo}[2]{#2}
\providecommand{\BIBentrySTDinterwordspacing}{\spaceskip=0pt\relax}
\providecommand{\BIBentryALTinterwordstretchfactor}{4}
\providecommand{\BIBentryALTinterwordspacing}{\spaceskip=\fontdimen2\font plus
\BIBentryALTinterwordstretchfactor\fontdimen3\font minus
  \fontdimen4\font\relax}
\providecommand{\BIBforeignlanguage}[2]{{%
\expandafter\ifx\csname l@#1\endcsname\relax
\typeout{** WARNING: IEEEtran.bst: No hyphenation pattern has been}%
\typeout{** loaded for the language `#1'. Using the pattern for}%
\typeout{** the default language instead.}%
\else
\language=\csname l@#1\endcsname
\fi
#2}}
\providecommand{\BIBdecl}{\relax}
\BIBdecl

\bibitem{lavie2017}
A.~Lavie, A.~Waibel, L.~S. Levin, M.~Finke, D.~Gates, M.~Gavald{\`{a}},
  T.~Zeppenfeld, and P.~Zhan, ``Janus-iii: speech-to-speech translation in
  multiple languages,'' in \emph{Proc. {ICASSP}}, 1997.

\bibitem{nakamura2006}
S.~Nakamura, K.~Markov, H.~Nakaiwa, G.~Kikui, H.~Kawai, T.~Jitsuhiro, J.~Zhang,
  H.~Yamamoto, E.~Sumita, and S.~Yamamoto, ``The {ATR} multilingual
  speech-to-speech translation system,'' \emph{{IEEE} Trans. Speech Audio
  Process.}, 2006.

\bibitem{sudoh}
R.~Fukuda, S.~Novitasari, Y.~Oka, Y.~Kano, Y.~Yano, Y.~Ko, H.~Tokuyama, K.~Doi,
  T.~Yanagita, S.~Sakti, K.~Sudoh, and S.~Nakamura, ``Simultaneous
  speech-to-speech translation system with transformer-based incremental asr,
  mt, and tts,'' in \emph{Proc. O-COCOSDA}, 2021.

\bibitem{tjandra2019}
A.~Tjandra, S.~Sakti, and S.~Nakamura, ``Speech-to-speech translation between
  untranscribed unknown languages,'' in \emph{Proc. {ASRU}}, 2019.

\bibitem{translatoron}
Y.~Jia, R.~J. Weiss, F.~Biadsy, W.~Macherey, M.~Johnson, Z.~Chen, and Y.~Wu,
  ``Direct speech-to-speech translation with a sequence-to-sequence model,'' in
  \emph{Proc. Interspeech}, 2019.

\bibitem{s2ut}
A.~Lee, P.-J. Chen, C.~Wang, J.~Gu, S.~Popuri, X.~Ma, A.~Polyak, Y.~Adi, Q.~He,
  Y.~Tang, J.~Pino, and W.-N. Hsu, ``Direct speech-to-speech translation with
  discrete units,'' in \emph{Proc. ACL}, 2022.

\bibitem{translatoron2}
Y.~Jia, M.~T. Ramanovich, T.~Remez, and R.~Pomerantz, ``Translatotron 2: Robust
  direct speech-to-speech translation,'' in \emph{Proc. ICML}, 2022.

\bibitem{ma-etal-2019-stacl}
M.~Ma, L.~Huang, H.~Xiong, R.~Zheng, K.~Liu, B.~Zheng, C.~Zhang, Z.~He, H.~Liu,
  X.~Li, H.~Wu, and H.~Wang, ``{STACL}: Simultaneous translation with implicit
  anticipation and controllable latency using prefix-to-prefix framework,'' in
  \emph{Proc. ACL}, 2019.

\bibitem{ma-etal-2020-simulmt}
X.~Ma, J.~Pino, and P.~Koehn, ``{S}imul{MT} to {S}imul{ST}: Adapting
  simultaneous text translation to end-to-end simultaneous speech
  translation,'' in \emph{Proc. AACL}, 2020.

\bibitem{ma-etal-2020-incremental}
M.~Ma, B.~Zheng, K.~Liu, R.~Zheng, H.~Liu, K.~Peng, K.~Church, and L.~Huang,
  ``Incremental text-to-speech synthesis with prefix-to-prefix framework,'' in
  \emph{Proc. EMNLP}, 2020.

\bibitem{stephenson21_interspeech}
B.~Stephenson, T.~Hueber, L.~Girin, and L.~Besacier, ``{Alternate Endings:
  Improving Prosody for Incremental Neural TTS with Predicted Future Text
  Input},'' in \emph{Proc. Interspeech}, 2021.

\bibitem{pseudolookahead}
T.~Saeki, S.~Takamichi, and H.~Saruwatari, ``Incremental text-to-speech
  synthesis using pseudo lookahead with large pretrained language model,''
  \emph{{IEEE} Signal Process. Lett.}, vol.~28, 2021.

\bibitem{pseudolookaheaddistilled}
T.~Saeki, S.~Takamichi, and Saruwatari, ``Low-latency incremental
  text-to-speech synthesis with distilled context prediction network,'' in
  \emph{Proc. ASRU}, 2021.

\bibitem{neuraliTTS}
T.~Yanagita, S.~Sakti, and S.~Nakamura, ``{Neural iTTS: Toward Synthesizing
  Speech in Real-time with End-to-end Neural Text-to-Speech Framework},'' in
  \emph{Proc. ISCA SSW}, 2019.

\bibitem{lookahead}
B.~Stephenson, L.~Besacier, L.~Girin, and T.~Hueber, ``What the future brings:
  Investigating the impact of lookahead for incremental neural {TTS},'' in
  \emph{Proc. Interspeech}, 2020.

\bibitem{RLiTTS}
D.~S.~R. Mohan, R.~Lenain, L.~Foglianti, T.~H. Teh, M.~Staib,
  A.~Torresquintero, and J.~Gao, ``Incremental text to speech for neural
  sequence-to-sequence models using reinforcement learning,'' in \emph{Proc.
  Interspeech}, 2020.

\bibitem{tacotron}
Y.~Wang, R.~J. Skerry{-}Ryan, D.~Stanton, Y.~Wu, R.~J. Weiss, N.~Jaitly,
  Z.~Yang, Y.~Xiao, Z.~Chen, S.~Bengio, Q.~V. Le, Y.~Agiomyrgiannakis,
  R.~Clark, and R.~A. Saurous, ``Tacotron: Towards end-to-end speech
  synthesis,'' in \emph{Proc. Interspeech}, 2017.

\bibitem{tacotron2}
J.~Shen, R.~Pang, R.~J. Weiss, M.~Schuster, N.~Jaitly, Z.~Yang, Z.~Chen,
  Y.~Zhang, Y.~Wang, R.~Ryan, R.~A. Saurous, Y.~Agiomyrgiannakis, and Y.~Wu,
  ``Natural {TTS} synthesis by conditioning wavenet on {M}el spectrogram
  predictions,'' in \emph{Proc. {ICASSP}}, 2018.

\bibitem{radford2019language}
A.~Radford, J.~Wu, R.~Child, D.~Luan, D.~Amodei, I.~Sutskever \emph{et~al.},
  ``Language models are unsupervised multitask learners,'' \emph{OpenAI blog},
  vol.~1, no.~8, p.~9, 2019.

\bibitem{fastspeech2}
Y.~Ren, C.~Hu, X.~Tan, T.~Qin, S.~Zhao, Z.~Zhao, and T.~Liu, ``Fastspeech 2:
  Fast and high-quality end-to-end text to speech,'' in \emph{Proc. ICLR},
  2021.

\bibitem{cooper1981segmental}
W.~E. Cooper and M.~Danly, ``Segmental and temporal aspects of utterance-final
  lengthening,'' \emph{Phonetica}, vol.~38, no. 1-3, pp. 106--115, 1981.

\bibitem{berkovits1993utterance}
R.~Berkovits, ``Utterance-final lengthening and the duration of final-stop
  closures,'' \emph{Journal of Phonetics}, vol.~21, no.~4, pp. 479--489, 1993.

\bibitem{fisher}
M.~Post, G.~Kumar, A.~Lopez, D.~G. Karakos, C.~Callison{-}Burch, and
  S.~Khudanpur, ``Improved speech-to-text translation with the fisher and
  callhome spanish-english speech translation corpus,'' in \emph{Proc. IWSLT},
  2013.

\bibitem{di-gangi-etal-2019-must}
M.~A. Di~Gangi, R.~Cattoni, L.~Bentivogli, M.~Negri, and M.~Turchi,
  ``{M}u{ST}-{C}: a {M}ultilingual {S}peech {T}ranslation {C}orpus,'' in
  \emph{Proc. NAACL}, 2019.

\bibitem{sennrich-etal-2016-neural}
R.~Sennrich, B.~Haddow, and A.~Birch, ``Neural machine translation of rare
  words with subword units,'' in \emph{Proc. ACL}, 2016.

\bibitem{kudo-richardson-2018-sentencepiece}
T.~Kudo and J.~Richardson, ``{S}entence{P}iece: A simple and language
  independent subword tokenizer and detokenizer for neural text processing,''
  in \emph{Proc. EMNLP}, 2018.

\bibitem{ljspeech17}
K.~Ito and L.~Johnson, ``The {LJ Speech} dataset,''
  \url{https://keithito.com/LJ-Speech-Dataset/}, 2017.

\bibitem{css10}
K.~Park and T.~Mulc, ``{CSS10:} {A} collection of single speaker speech
  datasets for 10 languages,'' in \emph{Proc. Interspeech}, 2019.

\bibitem{mfa}
M.~McAuliffe, M.~Socolof, S.~Mihuc, M.~Wagner, and M.~Sonderegger, ``Montreal
  forced aligner: Trainable text-speech alignment using kaldi,'' in \emph{Proc.
  Interspeech}, 2017.

\bibitem{schultz-schlippe-2014-globalphone}
T.~Schultz and T.~Schlippe, ``{G}lobal{P}hone: Pronunciation dictionaries in 20
  languages,'' in \emph{Proc LREC}, 2014.

\bibitem{wang-etal-2020-fairseq}
C.~Wang, Y.~Tang, X.~Ma, A.~Wu, D.~Okhonko, and J.~Pino, ``Fairseq {S}2{T}:
  Fast speech-to-text modeling with {Fairseq},'' in \emph{Proc. AACL}, 2020.

\bibitem{hifigan}
J.~Kong, J.~Kim, and J.~Bae, ``{HiFi-GAN}: Generative adversarial networks for
  efficient and high fidelity speech synthesis,'' in \emph{Proc. NeurIPS},
  2020.

\bibitem{Wave-Tacotron}
R.~J. Weiss, R.~J. Skerry{-}Ryan, E.~Battenberg, S.~Mariooryad, and D.~P.
  Kingma, ``{Wave-Tacotron}: Spectrogram-free end-to-end text-to-speech
  synthesis,'' in \emph{Proc. {ICASSP}}, 2021.

\bibitem{wang-etal-2021-fairseq}
C.~Wang, W.-N. Hsu, Y.~Adi, A.~Polyak, A.~Lee, P.-J. Chen, J.~Gu, and J.~Pino,
  ``fairseq s{\^{}}2: A scalable and integrable speech synthesis toolkit,'' in
  \emph{Proc. EMNLP}, 2021.

\bibitem{MCD}
J.~Kominek, T.~Schultz, and A.~W. Black, ``Synthesizer voice quality of new
  languages calibrated with mean mel cepstral distortion,'' in \emph{Proc.
  {SLTU}}, 2008.

\bibitem{wav2vec}
A.~Baevski, Y.~Zhou, A.~Mohamed, and M.~Auli, ``wav2vec 2.0: {A} framework for
  self-supervised learning of speech representations,'' in \emph{Proc.
  NeurIPS}, 2020.

\bibitem{xlsr53}
A.~Conneau, A.~Baevski, R.~Collobert, A.~Mohamed, and M.~Auli, ``Unsupervised
  cross-lingual representation learning for speech recognition,'' in
  \emph{Proc. Interspeech}, 2021.

\bibitem{ardila-etal-2020-common}
R.~Ardila, M.~Branson, K.~Davis, M.~Kohler, J.~Meyer, M.~Henretty, R.~Morais,
  L.~Saunders, F.~Tyers, and G.~Weber, ``Common voice: A massively-multilingual
  speech corpus,'' in \emph{Proc. LREC}, Marseille, France, 2020.

\bibitem{post-2018-call}
M.~Post, ``A call for clarity in reporting {BLEU} scores,'' in \emph{Proc.
  WMT}, 2018.

\bibitem{ma2022direct}
X.~Ma, H.~Gong, D.~Liu, A.~Lee, Y.~Tang, P.-J. Chen, W.-N. Hsu, P.~Koehn, and
  J.~Pino, ``Direct simultaneous speech-to-speech translation with variational
  monotonic multihead attention,'' \emph{CoRR}, vol. abs/2110.08250, 2022.

\end{thebibliography}
